\documentclass[draft]{agujournal2019}
\usepackage{url} 
\usepackage{lineno}

\usepackage{soul}

\usepackage{amsmath}
\usepackage{booktabs}
\usepackage{multirow}

\usepackage{amssymb}
\usepackage{graphicx}
\usepackage{pifont}
\usepackage{siunitx}
\usepackage{makecell}
\usepackage{xcolor}
\usepackage{caption}
\usepackage{ragged2e}
\usepackage[commandnameprefix=always,markup=default,authormarkup=none]{changes}

\justifying
\journalname{JGR: Machine Learning and Computation}

\begin{document}

\title{OceanForecastBench: A Benchmark Dataset for Data-Driven Global Ocean Forecasting}

\authors{Haoming Jia\affil{1}$^{*}$, Yi Han\affil{1}\thanks{These authors contributed equally to this work.}, Xiang Wang\affil{1}, Huizan Wang\affil{1}, Wei Wu\affil{2}, Jianming Zheng\affil{3}, Peikun Xiao\affil{1}}

\affiliation{1}{College of Meteorology and Oceanography, National University of Defense Technology,
	Changsha,
	410000, 
	Hunan,
	China}
\affiliation{2}{College of Computer Science and Technology, National University of Defense Technology,
	Changsha,
	410000, 
        Hunan,
	China}
\affiliation{3}{State Key Laboratory of Mathematical Engineering and Advanced Computing,
	Wuxi,
	214000, 
	Jiangsu,
	China}

\correspondingauthor{Xiang Wang}{xiangwangcn@nudt.edu.cn}

\begin{keypoints}
\item We contribute to AI for Earth by introducing an open-source benchmark for data-driven global ocean forecasting
\item We preprocess multi-source data into a ready-to-use dataset for the development of data-driven forecasting models and evaluation pipelines
\item We conduct a detailed analysis of 5 typical baseline models, highlighting their strengths and limitations to guide future improvements
\end{keypoints}

\begin{abstract}
Global ocean forecasting aims to predict key ocean variables such as temperature, salinity, and currents, which is essential for understanding and describing oceanic phenomena.
In recent years, data-driven deep learning-based ocean forecast models, such as XiHe, WenHai, LangYa and AI-GOMS, have demonstrated significant potential in capturing complex ocean dynamics and improving forecasting efficiency.
Despite these advancements, the absence of open-source, standardized benchmarks has led to inconsistent data usage and evaluation methods.
This gap hinders efficient model development, impedes fair performance comparison, and constrains interdisciplinary collaboration.
To address this challenge, we propose OceanForecastBench, a benchmark offering three core contributions:
(1) A high-quality global ocean reanalysis data over 28 years for model training, including 4 ocean variables across 23 depth levels and 4 sea surface variables. 
(2) A high-reliability satellite and in-situ observations for model evaluation, covering approximately 100 million locations in the global ocean.
(3) An evaluation pipeline and a comprehensive benchmark with 5 typical baseline models, leveraging observations to evaluate model performance from multiple perspectives.
OceanForecastBench represents the most comprehensive benchmarking framework currently available for data-driven ocean forecasting, offering an open-source platform for model development, evaluation, and comparison.
The dataset and code are publicly available at: \url{https://github.com/Ocean-Intelligent-Forecasting/OceanForecastBench}.
\end{abstract}

\section*{Plain Language Summary}
Ocean forecasts help us understand ocean currents, temperature, salinity and climate impacts days ahead.
Recent advances in AI-based models, such as XiHe and WenHai, have shown great potential in improving the accuracy and efficiency of ocean predictions.
However, the lack of a standardized benchmark for training and evaluating these models poses a major barrier to fair comparison and reproducibility.
Researchers currently rely on disparate datasets and evaluation methods, which hinders scientific progress and model development.
To address this, we constructed OceanForecastBench, a comprehensive and open-source benchmark that provides:
(1) High-quality training data over 28 years of global ocean reanalysis (temperature, currents, salinity etc.) at different depths.
(2) Reliable evaluation data which includes millions of real-world satellite and sensor observations to evaluate model accuracy.
(3) Standardized evaluation of 5 baseline models across key performance, enabling clear comparisons.

\section{Introduction}
Global ocean forecasting is fundamentally important for supporting ocean activities, such as ensuring the safety of maritime navigation, improving climate predictions, and enhancing the management of marine resources~\cite{meehl2021initialized,payne2022skilful}. 
Conventional physics-driven methods achieve ocean forecasting by solving physical partial differential equations (PDEs) using numerical simulations.
However, the chaotic nature of ocean makes solving PDEs highly complex and time-consuming.
Consequently, these methods often demand substantial computational resources and time.

Artificial Intelligence for Earth (AI4Earth) is revolutionizing ocean forecasting by enabling the data-driven discovery of complex oceanographic patterns.
Recent advancements in AI have demonstrated significant potential for ocean forecasting~\cite{fengewu2023,chen2023fuxi,chae2024prediction,chen2024short}. Data-driven AI ocean forecasting models, such as XiHe~\cite{wang_XiHe_}, WenHai~\cite{cui2025forecasting}, LangYa~\cite{yang2024langya}, and AI-GOMS~\cite{xiong_AIGOMS_2023} achieve faster speeds, more accurate forecasts, longer-term forecasting capability, and lower computational costs~\cite{choudhury2024bharatbench}.
However, progress in this field still remains limited, with only a few studies.
The main reason is the absence of a public benchmark that enables fair algorithm comparison and improves research reproducibility, which creates professional barriers within the field~\cite{kaltenborn2023climateset,dueben_Challenges_2022}.
Benchmarks are crucial for facilitating algorithm comparison and model optimization, as demonstrated by ImageNet~\cite{deng2009imagenet}, WeatherBench~\cite{rasp_WeatherBench_2020}, and GLUE~\cite{wang2018glue}. 
Therefore, establishing an open-source, unified benchmark for data-driven global ocean forecasting is essential to propel progress in this domain.

However, creating such a benchmark for ocean forecasting involves several significant challenges: 
(1) \textbf{Requirement for expert knowledge in ocean forecasting-specific benchmark.}
Developing a benchmark involves defining tasks, providing standardized datasets, and evaluating models. Domain-specific knowledge is crucial to identify key variables and avoid task flaws. Ocean forecasting integrates data from multiple sources, so expert knowledge is needed for proper data processing and standardization. 
Additionally, using only the reanalysis data for model evaluation may introduce the risk of bias and reduced accuracy. Observations are often used to assess model performance, highlighting the need for an observation-based evaluation pipeline.
(2) \textbf{Complexity of processing multi-source data when constructing benchmark datasets.} Deep learning models require large datasets. Ocean forecasting relies on various data types (e.g., reanalysis, in-situ observations, and satellite observations) for accuracy. Global reanalysis data is ideal for training, offering consistency and comprehensive spatial coverage. Discrete observations, which record the ocean's true state, are suitable for model evaluation but come with challenges like inconsistent quality (e.g., noise interference) and format. Integrating these data sources requires careful selection and processing.
(3) \textbf{High time and computational resource demand in baseline acquisition}. 
Identifying appropriate baseline models is challenging, as few AI models exist for ocean forecasting. Adapting models from other fields requires additional efforts. Training models, especially deep learning ones, is computationally intensive. For instance, training a single SwinTransformer variant on an 8-GPU cluster takes around 30 hours, and hyperparameter optimization further increases this burden. Additionally, comparing results is time-consuming and requires expertise in oceanography.

This paper proposes a benchmark dataset for data-driven global ocean forecasting.
The task of the benchmark is to forecast key ocean variables, such as sea level anomaly (SLA), sea surface temperature (SST), zonal and meridional components of currents (Uo and Vo), temperature, and salinity.
Specifically, \textbf{for the training dataset}, we primarily use GLORYS12 (Global Ocean Physics Reanalysis)~\cite{jean-michel_Copernicus_2021}, one of the best products currently available.
To account for the influence of sea surface environmental factors, we also integrate sea surface wind from ERA5 (Fifth Generation Global Atmospheric Reanalysis)~\cite{bell2021era5} and SST data from OSTIA (Operational Sea Surface Temperature and Ice Analysis)~\cite{good2020current} into the training dataset.
The training dataset has been standardized to a uniform spatial and temporal resolution to ensure usability, derived from over 13TB of raw data. 
\textbf{For the evaluation dataset}, we utilize a range of observations (such as satellite remote sensing, in-situ observations, etc.) as the ground truth for evaluating model performance.
Specifically, we select the quality-controlled subsurface ocean temperature and salinity data from EN4 data~\cite{good2013en4,gouretski2010depth}, 6-hour interpolated SST and currents data from Global Drifter Program (GDP)~\cite{lumpkin2019global}, and SLA data from Global Ocean L3 Significant Wave Height From Reprocessed Satellite Measurements provided by Copernicus Marine Environment Monitoring Service (CMEMS L3). The evaluation dataset covers approximately 100 million observation locations in the global ocean. A standardized evaluation pipeline is provided to align forecast data with discrete observations and calculate evaluation metrics.
\textbf{For the baseline models}, we conduct a detailed evaluation of 5 representative models, including the operational numerical forecasting systems, and the advanced deep learning models. 
A detailed analysis of the evaluation results highlights the strengths and weaknesses of various baseline models, offering critical insights for future improvements.
OceanForecastBench standardizes both model training and evaluation processes, significantly simplifying data processing workflows.
This benchmark facilitates participation from scholars across diverse fields in ocean forecasting.
\begin{figure*}[!b]
\centering
\includegraphics[width=1\textwidth]{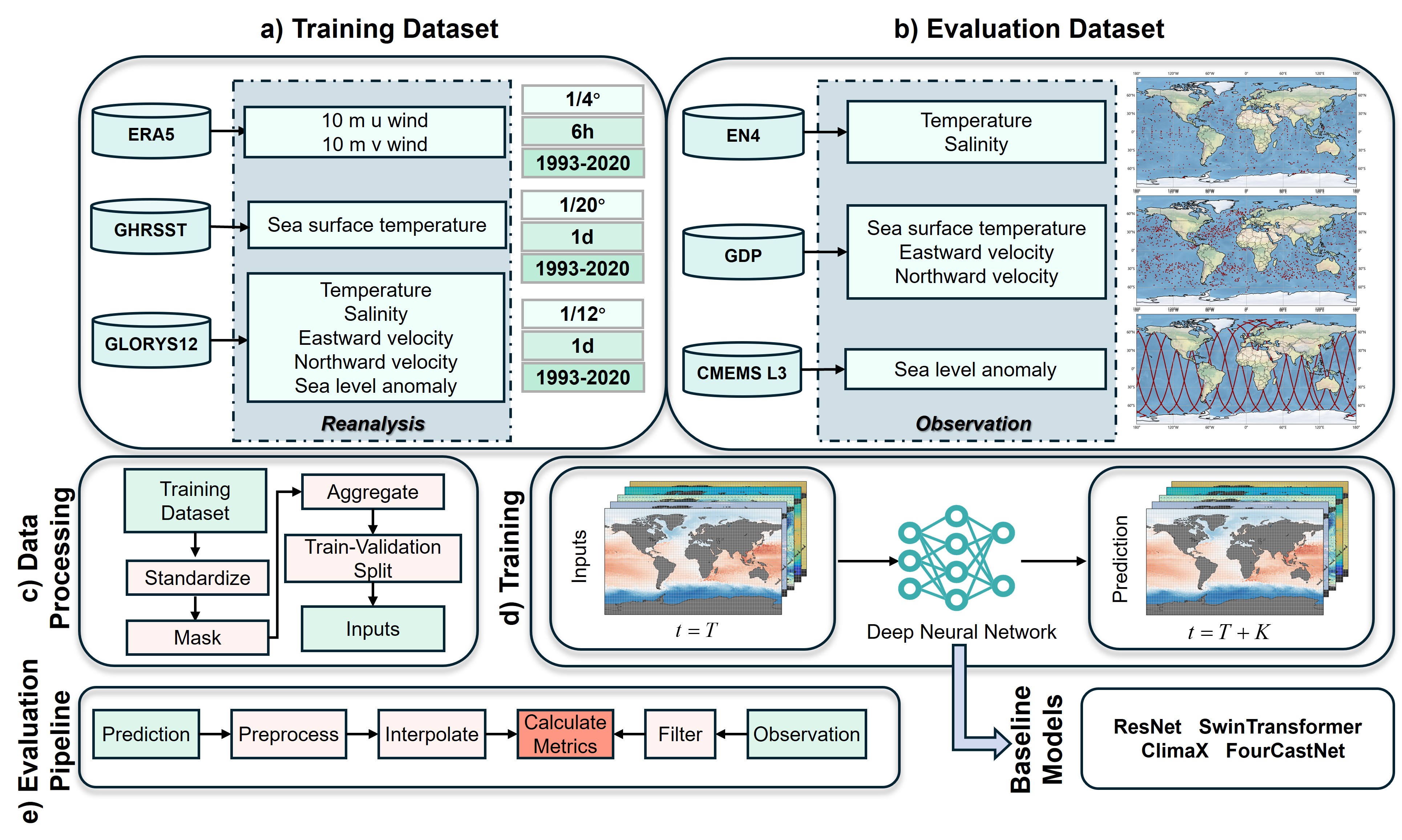}
\caption{Schematic of the OceanForecastBench data collection and processing pipeline. }
\label{fig:pipeline}
\end{figure*}
\section{The OceanForecastBench Dataset}
\label{datainfo} 
Figure~\ref{fig:pipeline} presents the schematic of the OceanForecastBench data collection and processing pipeline.
The OceanForecastBench dataset covers a wide range of ocean variables above 643.57m, including temperature, salinity, and currents, providing comprehensive data for ocean forecasting. 
It also integrates multi-source data, including reanalysis, satellite remote sensing, buoy observations, and drifter observations.
The detailed information about the datasets included in OceanForecastBench is as shown in Table \ref{reanalysis}.

\begin{table}[!b]
\centering
\setlength{\abovecaptionskip}{0pt}
\setlength{\belowcaptionskip}{5pt}
\caption{Detailed information about the data included in OceanForecastBench.
}
\resizebox{\textwidth}{!}{
	\begin{tabular}{c|c|c|c|c|c}
		\toprule
		\textbf{Data} & \textbf{Ocean Variables} & \textbf{Abbreviations}  &   \textbf{Unit} & \textbf{Levels} & \textbf{Source}\\ [0.5ex]
		\hline\hline

		\multirow{6}{*}{GLORYS12}&Sea surface height  & SSH & m & 1&\multirow{9}{*}{Numerical Model Reanalysis/Analysis}\\
		
		&Temperature  & T & \textcelsius & 50&\\
		&Salinity  & S & \text{PSU} & 50&\\
		&Eastward currents velocity  & Uo & $\text{m}~\text{s}^{-1}$ & 50&\\
		&Northward currents velocity  & Vo & $\text{m~s}^{-1}$ & 50&\\
		&Mean dynamic topography  & mdt & \unit{\meter} & 1 &\\
		\cline{1-5}
		\multirow{2}{*}{ERA5}&10 metre U wind component  & u10 & $\text{m~s}^{-1}$ & 1&\\
		&10 metre V wind component  & v10 & $\text{m~s}^{-1}$ & 1&\\
		\cline{1-5}
		OSTIA& Sea surface temperature  & SST & \unit{\kelvin} & 1&\\ \hline
		\multirow{2}{*}{EN4}&Temperature &T & \textcelsius & - & \multirow{5}{*}{In-situ Observation}\\ 
		&Salinity  &S&PSU&-&\\ \cline{1-5}
		\multirow{3}{*}{GDP}& Sea surface temperature & SST & \textcelsius & 1&\\ 
		& 15 m eastward velocity & Uo &  $\text{m~s}^{-1}$ & 1& \\ 
		& 15 m northward velocity & Vo &  $\text{m~s}^{-1}$  & 1& \\ \hline
		CMEMS L3 & Sea surface height & SSH & m & 1&Satellite Observation \\ 
		
		\bottomrule
	\end{tabular}
}
\label{reanalysis}
\end{table}

\subsection{Training Dataset}
\textbf{GLORYS12} reanalysis is a high-quality, long-time ocean variables data provided by CMEMS. 
GLORYS12 offers continuous estimates of the global ocean from 1993 to 2021, with a horizontal resolution of 0.083° (approximately 8 kilometers at the equator). 
The vertical structure contains 50 vertical layers extending from the sea surface to a depth of 5,000 meters, fully depicting the vertical structure of the ocean. 
GLORYS12 has been integrated into the OceanForecastBench training dataset to provide globally scaled data for ocean temperature, salinity, and currents, which are fundamental and essential ocean variables.
Mean dynamic topography (MDT) is also provided to facilitate the conversion of sea surface height (SSH) and SLA. 
The total size of the GLORYS12 data used in OceanForecastBench is approximately 13TB.

\textbf{ERA5} is a global meteorological reanalysis data provided by the European Centre for Medium-Range Weather Forecasts (ECMWF). 
ERA5 offers high-resolution sea surface wind data ($0.25^{\circ}\times0.25^{\circ}$) with a temporal resolution of once per six hours. 
Sea surface wind data is a primary external force driving ocean circulation. 
So ERA5 has been integrated into the training dataset to enable models to more accurately describe the impact of sea surface wind and predict currents velocity.
The total size of ERA5 data used is approximately 170GB.

\textbf{OSTIA} is a high-resolution, global-scale data product developed to provide accurate and timely information on SST and sea ice concentration. 
OSTIA integrates satellite observations from multiple sensors and numerical model forecasts to produce daily analyses at a spatial resolution of approximately 1/20° (6 km).
SST is a crucial indicator of the energy exchange between the ocean and the atmosphere. 
OSTIA has been integrated into the training dataset to allow models to better simulate physical processes in the ocean surface layer.
The total size of the OSTIA data used is approximately 160GB.

\subsection{Evaluation Dataset}
\textbf{EN4} observation is a high-quality controlled data of global ocean temperature and salinity, covering the time range from 1900 to the 2025. 
The data primarily originates from the global Argo buoy network, which consists of about 3,000 profiling buoys. 
These data undergo a rigorous quality control process, including outlier detection, consistency checks, and comparison with historical data, to ensure their accuracy and reliability. 
EN4 has been integrated into the evaluation dataset for comparing the performance of ocean temperature and salinity forecasts. 

\textbf{GDP} observation is based on data from ocean surface drifters, which have undergone rigorous quality control and interpolation at 6-hour intervals to provide a continuous record of SST and currents. 
With high temporal resolution and global coverage, the GDP data serves as ground truth for evaluating model predictive capabilities of SST and currents velocity. 
GDP data has been integrated into the evaluation dataset for comparing the performance of SST and currents forecasts. 

\textbf{CMEMS L3} provides a high-precision, long-time series of SLA remote sensing data. It provides a long-term, high-resolution sea level anomaly (SLA) record since 1993 with a sampling rate of 1 Hz (approximately 7 km resolution). 
The data has been accurately corrected for the atmosphere and considers the effects of ocean tides.  
CMEMS L3 has been integrated into the evaluation dataset for comparing the performance of SLA forecasts. 
The total evaluation dataset data size is 57GB.

\subsection{Data Processing}

This benchmark provides a standardized data processing workflow that extracts specified data from the original files according to task requirements, processes the data, and outputs high-dimensional arrays for model training.
Figure~\ref{fig:pipeline}c illustrates the flow of data processing.

The training data was collected from three sources (with different variables and resolutions) necessitating standardization to a uniform resolution.
Since the original GLORYS12 data was substantial, containing daily 0.083° gridded data with 50 vertical levels from 1993 to 2020, the entire data size amounted to approximately 13TB. 
Considering the dataset's volume, GPU memory limitations, and available computing resources of most researchers, it was necessary to downsample both the spatial resolutions and the number of vertical levels to reduce the hardware demands~\cite{rasp2020weatherbench}.
However, excessive downsampling can result in significant information loss and  the introduction of errors, which could affect model prediction accuracy~\cite{bremnes2024evaluation}.
After careful analysis and experiments, a spatial resolution of 1.40625° and 23 vertical levels were chosen as a balanced trade-off between resource limitations and the preservation of data fidelity.
The chosen input variables and vertical levels are as described in Table \ref{variables_description}.
\begin{table}[!b]
\setlength{\abovecaptionskip}{0pt}
\setlength{\belowcaptionskip}{5pt}
\caption{ Ocean variables and depths modeled by OceanForecastBench. The input variables include 4 sea surface variables and 4 ocean variables with 23 vertical layers.}
\resizebox{\linewidth}{!}{%
	\begin{tabular}{c|m{0.12\linewidth}<{\centering}m{0.12\linewidth}<{\centering}m{0.12\linewidth}<{\centering}m{0.12\linewidth}<{\centering}m{0.12\linewidth}<{\centering}}
		\hline
		\textbf{Sea surface variables (4)} & \multicolumn{5}{m{0.7\linewidth}<{\centering}}{SST, SLA, zonal component of sea surface 10m wind, meridional component of sea surface 10m wind} \\ \hline
		\textbf{\begin{tabular}[c]{@{}c@{}}Ocean variables\\ (4*23)\end{tabular}} & \multicolumn{5}{m{0.7\linewidth}<{\centering}}{temperature, salinity, zonal component of currents velocity, meridional component of currents velocity} \\ \hline
		\multirow{5}{*}{\textbf{\begin{tabular}[c]{@{}c@{}}Depths of vertical levels \\ (23)\end{tabular}}} & 0.49m & 2.65m & 5.08m & 7.93m & 11.41m \\
		& 15.81m & 21.60m & 29.44m & 40.34m & 55.76m \\
		& 77.85m & 92.32m & 109.73m & 130.67m & 155.85m \\
		& 186.13m & 222.48m & 266.04m & 318.13m & 380.21m \\
		& 453.94m & 541.09m & 643.57m &  &  \\ \hline
	\end{tabular}%
}
\label{variables_description}
\end{table}

\textbf{Standardize.}
The Climate Data Operators were used to regrid the three data sources to a uniform resolution of 1.40625° over the same latitude and longitude range of $80^{\circ}\text{S}-90^{\circ}\text{N}$ and $180^{\circ}\text{W}-180^{\circ}\text{E}$, resulting in a grid size of $121\times256$.
Specially ERA5 is available at 4 time points each day: 0:00, 6:00, 12:00, and 18:00. 
Data from these four points is averaged to align the temporal resolution. 

\textbf{Mask.}
The proportion of ocean and land on earth is 70.8\% and 29.2\% respectively.
So we use an ocean-land mask file extracted from GLORYS12 to distinguish between ocean and land areas.
All land positions are filled with zero values before inputting into the model.

\textbf{Aggregate.}
The three data sources are combined according to the specified variable order to produce a standardized, deep learning-ready data format. 
The single-layer variables of SLA, 10m wind field, SST, and multi-layer temperature, salinity, and currents are combined into a multidimensional grid array.

\textbf{Train-Validation Split.}
The dataset is divided into training and validation datasets. (a) The data from 1993 to 2017 is used to train the deep learning model, and (b) the data from 2018 to 2020 is used as a validation set to validate the model performance during the model training phase.

\subsection{Evaluation Pipeline}
Figure~\ref{fig:pipeline}e illustrates the steps involved in calculating the metrics.
In this work, we have adopted a standardized data processing workflow for tensor-type data output by deep learning models to ensure data consistency and accessibility. 
Specifically, through a data processing script, the high-dimensional data in the array is split and extracted into multiple independent files, each of which stores a specific ocean variable.

A core step in model performance evaluation is accurately aligning the grid forecasts generated by the deep learning model with discrete observations in both the time and spatial dimensions.
The evaluation pipeline rigorously filters daily observations based on quality identifiers to identify data points available for evaluation.

\section{Problem Definition}
\label{des}
Global ocean forecasting predicts ocean variables supporting decision-making in ocean-related activities. 
Within the OceanForecastBench framework, deep learning models are used to forecast global ocean states for the next 1 to $k$ days.

Mathematically, the ocean forecasting task is a multivariate input-output mapping regression problem. The input data is represented as a three-dimensional tensor $X^{t}\in\mathbb{R}^{{C_{in}}\times{H}\times{W}}$, 
where $t$ denotes the current time step, $C_{in}=96$ indicates the total number of input ocean variables:  4 sea surface variables, and 4 ocean variables with 23 layers (see Table~\ref{variables_description}). $H=121$ and $W=256$ represent the number of grid points in the latitude and longitude directions, respectively.
The forecasting models aim to produce \textit{k}-step-ahead predictions $\hat{X}^{t+1:t+k}=(\hat{X}^{t+1},...,\hat{X}^{t+k})$, where $\hat{X}^{t+\Delta 
t}\in\mathbb{R}^{{C_{out}}\times{H}\times{W}},\Delta t=1,..,k.$
Here, $C_{out}=94$ indicates that the output variables excluding the zonal and meridional wind components used as the atmospheric forcing.

To evaluate the models, the forecast results need to be mapped from the forecast space $\hat{Y}_{fct}(t,pos)$ to the observation space $\hat{Y}_{obs}(t,pos)$. 
The mapping operator $\mathcal{M}$ interpolates the forecast data from a regular grid to the discrete observation locations:
\begin{equation}
\mathcal{M}:\hat{Y}_{fct}(t,pos)\to \hat{Y}_{obs}(t,pos)
\end{equation}
where $pos=[\mathrm{Longitude, Latitude}]^\top\in\mathbb{R}^{2},t=[\mathrm{Time}]\in\mathbb{R}^+$,  "$\to$" indicates time resolution alignment and spatial position alignment.

The evaluation metrics are then computed as:
\begin{equation}
\textsc{Metrics}=\textsc{Eval}(\hat{Y}_{obs}(t,pos),Y_{obs}(t,pos))
\end{equation}
where $\textsc{Eval}$ denotes a series of operations used to calculate the metrics (see Section \ref{metrics_intro}), $Y_{obs}(t,pos)$ represents the observations from the evaluation dataset.

\section{Experiments}
We conducted extensive experiments using the OceanForecastBench dataset to evaluate and compare the performance of various baseline models. The forecasting lead times are set from 1 to 10 days.
We use the Operational Mercator global ocean analysis from 2022 to 2023 as the initial conditions for the model forecast.
Leveraging evaluation data from 2022 to 2023, we performed a comprehensive and in-depth analysis of the experimental results from multiple perspectives.
These experiments were implemented using the PyTorch framework and performed on 8 NVIDIA RTX 4090 GPU.

\subsection{Benchmark Models}
OceanForecastBench selected and evaluated baseline models from numerical methods and deep learning methods:

The numerical forecast model chosen for this study is the French Physical System (PSY4), which is recognized as one of the most advanced operational forecasting systems. PSY4 baseline is obtained by downsampling the high-resolution operational forecasts which is obtained from \url{https://doi.org/10.48670/moi-00016}. 
The deep learning-based forecast models, including ResNet, SwinTransformer, ClimaX, and FourCastNet, are trained from scratch using the training dataset. Details of their hyperparameters are provided in Table \ref{baselines}.

\textbf{PSY4} is a 0.083° global operational forecast system developed by Mercator Ocean International. It is based on traditional numerical forecasting methods, that achieve forecasting by solving the partial differential equations of the physical models using supercomputers~\cite{lellouche2018mercator}.
To maintain consistency with other baseline models, the resolution of PSY4's forecasts was downsampled from 0.083° to 1.40625° during evaluation.

\textbf{ResNet} (Residual Network) is a deep neural network architecture that introduces residual connections to facilitate the training of very deep models. By allowing information to bypass certain layers via skip connections, ResNet enables the network to learn residual functions rather than direct mappings. 
This design mitigates vanishing gradient issues and reduces performance degradation, thereby improving training stability in deep architectures~\cite{he2016deep}.

\textbf{SwinTransformer} is a visual model utilizing shifted-window self-attention to improve for computational efficiency. Its hierarchical structure progressively merges patches for multi-resolution feature maps, enabling multi-resolution feature extraction for both local and global information, making it ideal for high-resolution tasks~\cite{liu_Swin_2021}.

\textbf{ClimaX} is a flexible and generalizable pretrained deep learning model for a wide range of climate and weather-related tasks. It employs a Vision Transformer (ViT) architecture to process spatiotemporal climate data and supports multidimensional tensor-to-tensor conversions~\cite{nguyen2023climax}.
Pretrained on global climate datasets using a self-supervised learning objective, ClimaX can be fine-tuned for various downstream weather tasks.

\textbf{FourCastNet} is a data-driven weather forecasting model based on ViT and Adaptive Fourier Neural Operators (AFNO).
By replacing traditional numerical solvers with a learned operator, it enables rapid prediction of the evolution of global atmospheric variables.
Trained on decades of reanalysis data, FourCastNet achieves competitive accuracy with traditional numerical weather forecasting models while offering orders-of-magnitude faster inference, making it highly suitable for real-time and large-scale forecasting applications~\cite{kurth_FourCastNet_2023}.

\begin{table}[h]
\centering
\setlength{\abovecaptionskip}{0pt}
\setlength{\belowcaptionskip}{5pt}
\caption{ Summary of baselines.}
\resizebox{\textwidth}{!}{
	\begin{tabular}{m{0.2\textwidth}<{\centering}|m{0.4\textwidth}<{\centering}|m{0.1\textwidth}<{\centering}|m{0.4\textwidth}<{\centering}}
		\toprule
		\textbf{Baselines}  & \textbf{Architecture} & \textbf{Parmas} & \textbf{Parameter Configuration}\\ \hline\hline
		
		PSY4 & Navier-Stokes equations & - & -\\ \hline
		ResNet & Convolutional neural network, Residual block & 51.4M & dropout:0.1, learning rate:$5e-4$, optimizer: AdamW, weight decay: $1e-5$\\ \hline
		SwinTransformer & Hierarchical feature map, Shift window mechanism & 38.2M & dropout: $0.1$, drop path: $0.2$, attn drop: $0$, learning rate:$5e-5$, optimizer: AdamW, weight decay: $1e-5$ \\ \hline 
		ClimaX & Variable tokenization, Variable aggregation, ViT & 65.7M & dropout: $0.1$, drop path: $0.2$, learning rate: $5e-4$, optimizer: AdamW, weight decay: $1e-5$\\ \hline
		FourCastNet & Adaptive fourier neural operators, ViT & 20.9M & dropout: $0$, learning rate: $1e-5$, optimizer: Adam\\ 
		
		\bottomrule
	\end{tabular}
}
\label{baselines}
\end{table}

\subsection{Evaluation Metrics}
\label{metrics_intro}
The evaluation metrics are calculated individually for each ocean variable of the forecast. 
Four metrics are employed: Root Mean Squared Error (RMSE), Bias, Anomaly Correlation Coefficient (ACC), and Climatology Skill Score (CSS). 
For the evaluation at time $t$,  let $F^{t}_{i}$, $O^{t}_{i}$, and $C^{t}_{i}$ represent the predicted value, the observed value, and the climatological mean value at each grid point $i$, respectively.
RMSE is calculated as follows:
\begin{equation}
\text{RMSE}=\sqrt{\frac{1}{N}\sum_{i=1}^{N}(F^{t}_{i}-O^{t}_{i})^{2}}
\end{equation}
Bias is calculated as follows:
\begin{equation}
\text{Bias}=\frac{1}{N}\sum_{i=1}^N(F^{t}_{i}-O^{t}_{i})
\end{equation}
ACC is calculated as follows:
\begin{equation}
\text{ACC}=\frac{\sum_{i=1}^N(F^{t}_{i}-C^{t}_{i})(O^{t}_{i}-C^{t}_{i})}{\sqrt{\sum_{i=1}^N(F^{t}_{i}-C^{t}_{i})^2}\sqrt{\sum_{i=1}^N(O^{t}_{i}-C^{t}_{i})^2}}
\end{equation}
In addition to the above commonly used metrics, the OceanForecastBench also utilizes  the CSS metric~\cite{ryan_GODAE_2015}. CSS quantifies the relative improvement of a forecast system compared to the long-term climatological average. A positive CSS value indicates that the forecast system outperforms the climate average. 
In contrast, a negative CSS means the forecast system underperforms relative to the climate average. 
CSS is computed using the formula:
\begin{equation}
	\text{CSS}=1-\frac{\text{RMSE}_{\text{Forecast}}}{\text{RMSE}_{\text{Climatology}}}
\end{equation}
\subsection{Overall Forecasting Performance}

\begin{figure}
\centering
\includegraphics[width=1\textwidth]{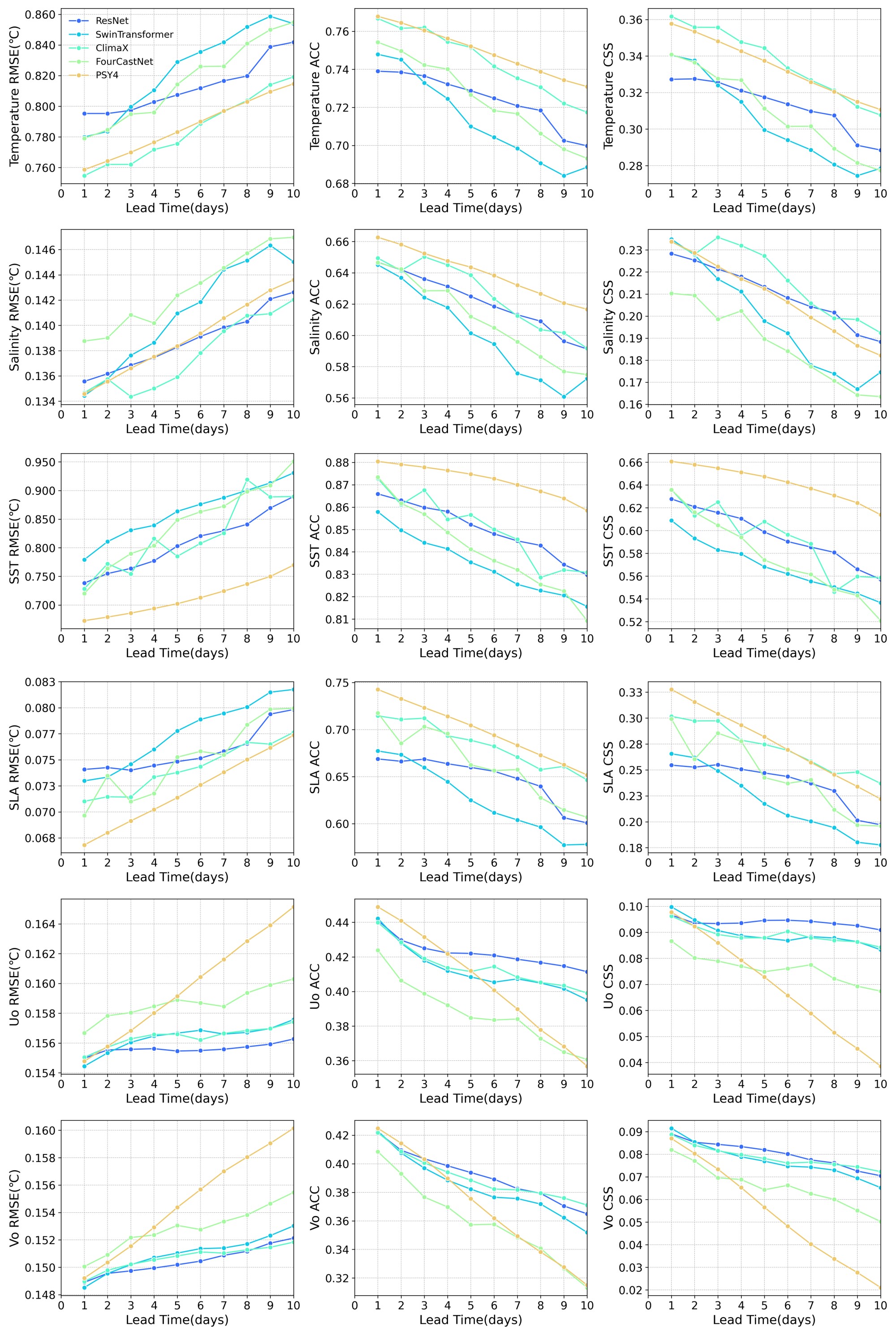}
\caption{The RMSE (lower is better), ACC (higher is better), and CSS (higher is better) of the temperature, salinity, SST, SLA, Uo and Vo for various baselines, over a forecast period ranging from 1 to 10 days. RMSE, ACC, and CSS are computed using observations provided by OceanForecastBench.}
\label{gmetrics}
\end{figure}

Figure~\ref{gmetrics} presents an assessment of the model's ability to provide forecasts of six key ocean variables over lead times ranging from 1 to 10 days.
The models were evaluated based on observations from 2022 to 2023.
For each lead time, the RMSE, Bias, ACC, and CSS were averaged globally from January 1, 2022, to December 31, 2023.
To enable unified evaluation based on OceanForecastBench, this study follows the international Ocean Model Intercomparison Project (OMIP) standard procedure \cite{griffies2016omip, orr2017biogeochemical}, in which raw forecast outputs of PSY4 are spatially downsampled to a standardized resolution of OceanForecastBench (1.40625°). 

As illustrated in Figure~\ref{gmetrics}, the RMSE values for all models generally increase with lead time, indicating a degradation in prediction accuracy as the forecast horizon extends.
The ACC and CSS metrics similarly exhibit a downward trend, further reinforcing this observation. 
Notably, the CSS valuess remain positive for all models even at longer lead times, indicating that the models retain some predictive skill beyond climatological predictions.
Another key observation is that PSY4 performs well at short lead times, but its forecast errors grow significantly as lead time increases. 
For example, as shown in Figure~\ref{gmetrics}, the RMSE of the SLA forecast at a 1-day lead time reveals a significant difference for PSY4 compared to other baseline models. However, between the 1-day and 6-day lead times, the RMSE of PSY4 gradually converges with those of the other models.
This is likely due to the model's use of a time-stepping mechanism to progressively advance the forecast. At each step, the model computes the next state based on the current state, leading to the gradual accumulation of errors~\cite{khaki20212019}. 
In contrast, the deep learning models in this study are one-step prediction strategy, directly learning the mapping from input to output data. This reduces the propagation and compounding of errors during forecasting~\cite{tang2025predicting}. 

For temperature and salinity prediction, the RMSE for the ClimaX remains relatively low compared to other models, indicating its superior performance in forecasting subsurface ocean variables.  
In the case of SST prediction, the PSY4 demonstrates superior performance, maintaining lower RMSE, higher ACC and CSS values compared to other models, especially at longer lead times.
The results can be attributed to its foundation in physical equations, which describe the evolution of SST.
Compared to ocean temperatures, the controlled processes of SST are relatively simpler, as they are primarily influenced by heat exchange with the atmosphere~\cite{wang2025advancing}. 
Therefore, the numerical model structure and parameterization processes of SST are relatively easier to fit and optimize.
For SLA prediction, the PSY4 shows the best performance before the 6-day forecast lead time.
The evaluation of currents velocity is performed by vertically interpolating the model forecasts to match the observed depth.
As shown in the Figure~\ref{gmetrics}, all deep learning models outperform the numerical model in long lead time currents velocity forecast.
While the numerical model shows the steepest increase in RMSE with increasing lead time, indicating its relatively poor performance at longer lead times.
Furthermore, we also observed that during the 1-10 day forecast period, the ACC of velocity forecasts is generally below 0.6, which is typically considered unreliable. This indicates that the reliability of various models in currents velocity forecasts still needs improvement.
\begin{figure*}[!t]
\centering
\includegraphics[width=1\textwidth]{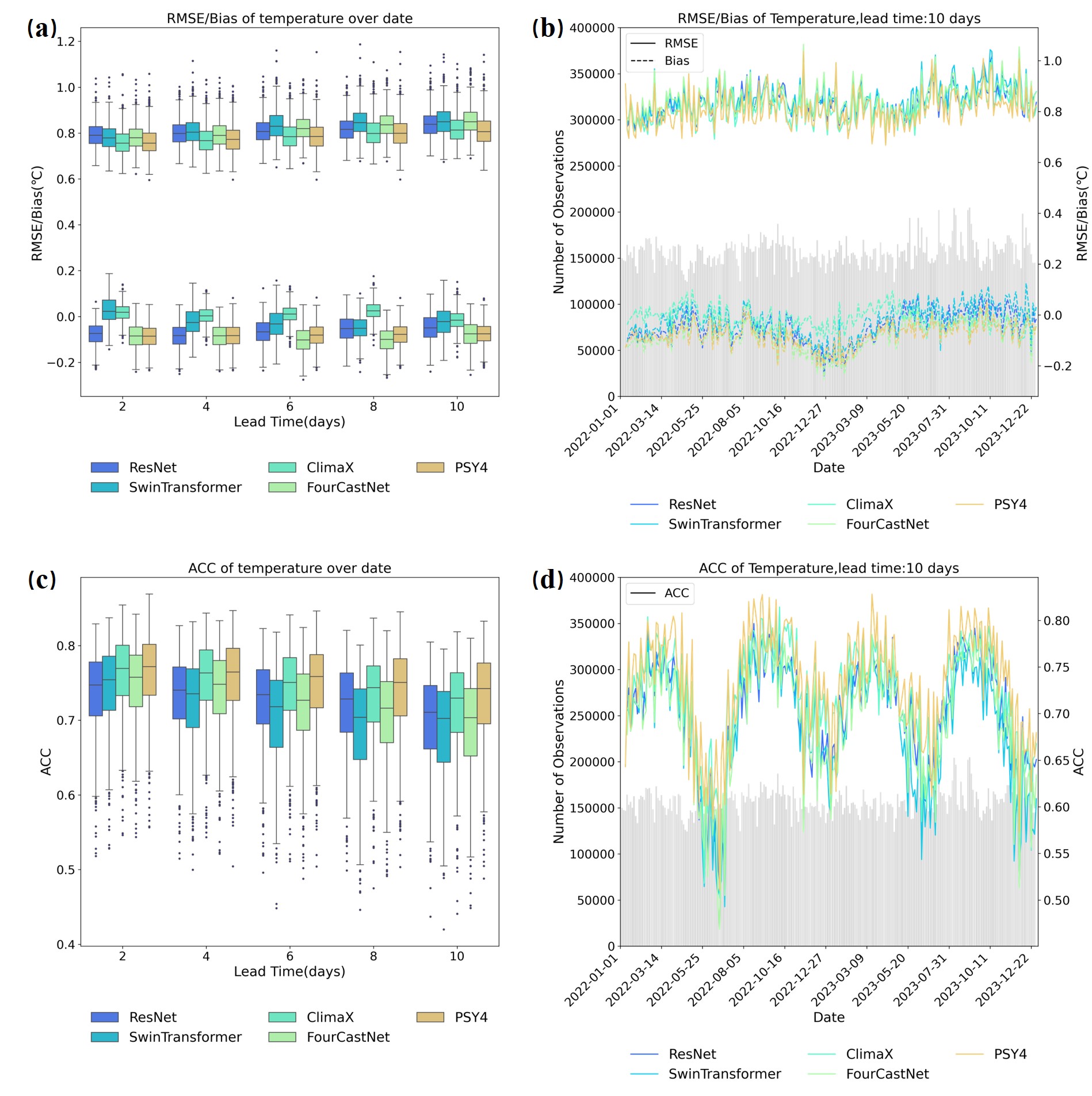} 
\caption{Forecast accuracy against temperature observations taken by EN4. (a) RMSE/Bias as a function of forecast lead time. The boxes are the interquartile range and the 75th percentile. (b) RMSE/Bias of 10-day forecasts as a function of date. (c) Similar to (a) with ACC as the metric. (d) Similar to (b) with ACC as the metric.}
\label{temperature_date} 
\end{figure*}

\subsection{Forecasting Performance in Various Aspects}
\textbf{Daily forecasting performance.}
The performance of temperature forecasts was assessed using three key metrics: RMSE, Bias, and ACC across lead times of 1 to 10 days.
Figure~\ref{temperature_date} shows the evaluation results of models by date.
Figure~\ref{temperature_date}a presents the RMSE and Bias distributions for all models across lead times.
As lead time increases, RMSE values for all models exhibit a consistent upward trend, reflecting the inherent challenges of long lead time forecasting.
Notably, PSY4 demonstrate lower RMSE values compared to other baselines, particularly at longer lead times (e.g., 10-day forecasts).
Figure~\ref{temperature_date}c illustrates the ACC values for all models across lead times. 
As the forecast lead time increases, the ACC value distribution of the PSY4 is more stable compared to other baseline models, highlighting its ability to maintain forecast accuracy over extended periods of temperature forecasting.
As shown in Figure~\ref{temperature_date}d, the ACC metric fluctuates markedly, with periods of high correlation (ACC $> 0.75$) interspersed with phases of low correlation (ACC $< 0.65$). 
These variations may be attributed to the dynamic nature of temperature anomalies.

\begin{figure*}[!t]
\centering
\includegraphics[width=1\textwidth]{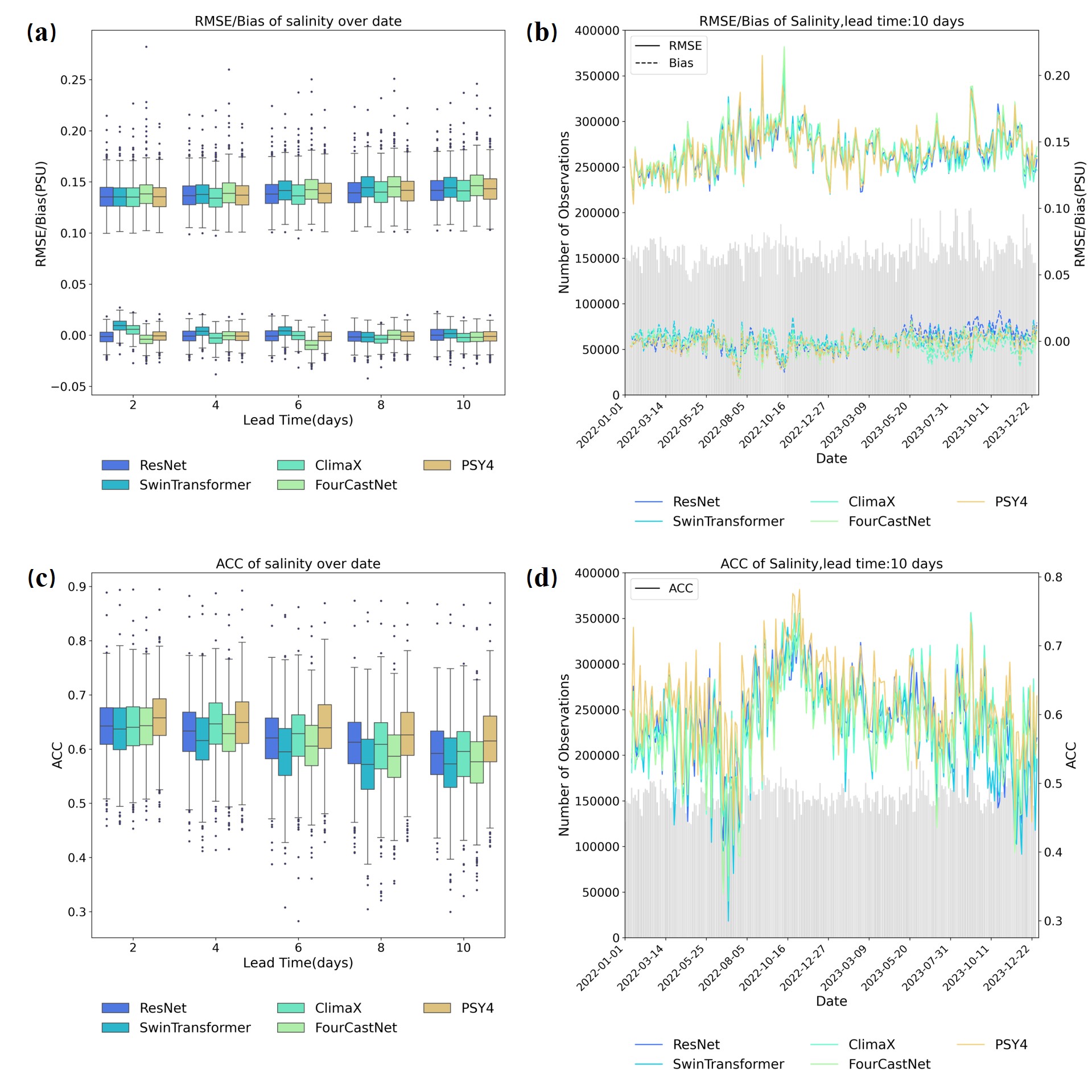} 
\caption{Forecast accuracy against salinity observations taken by EN4. (a) RMSE/Bias as a function of forecast lead time. The boxes are the interquartile range and the 75th percentile. (b) RMSE/Bias of 10-day forecasts as a function of date. (c) Similar to (a) with ACC as the metric. (d) Similar to (b) with ACC as the metric.}
\label{salinity_date} 
\end{figure*}
\begin{figure*}[!t]
\centering
\includegraphics[width=1\textwidth]{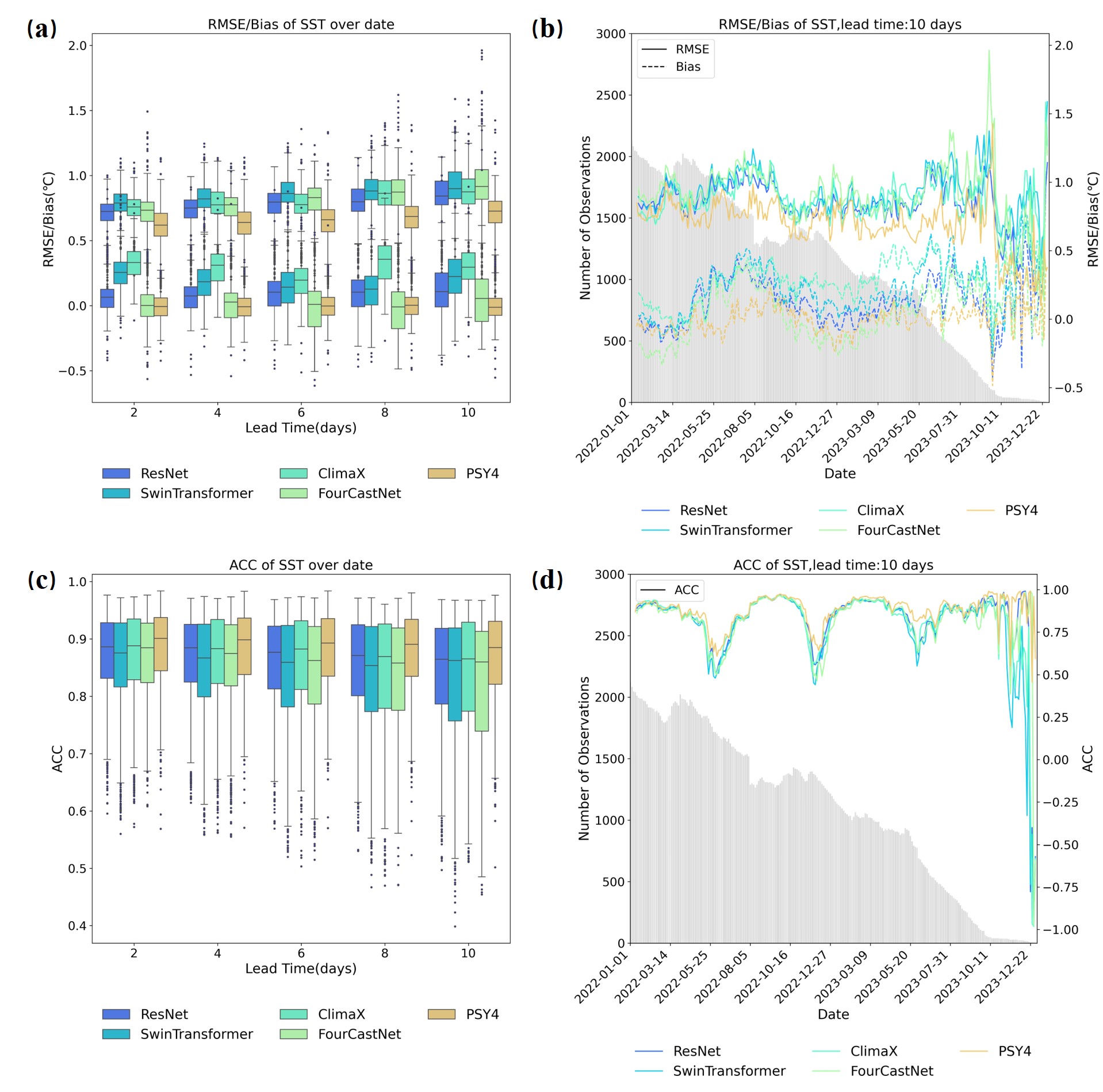} 
\caption{Forecast accuracy against SST observations taken by GDP. (a) RMSE/Bias as a function of forecast lead time. The boxes are the interquartile range and the 75th percentile. (b) RMSE/Bias of 10-day forecasts as a function of date. (c) Similar to (a) with ACC as the metric. (d) Similar to (b) with ACC as the metric.}
\label{sst_date} 
\end{figure*}

The performance of various models in predicting salinity are summarized in the Figure~\ref{salinity_date}.
The RMSE of salinity forecasts showed a slow increasing trend with increasing lead time, indicating that the models showed high stability in forecasting salinity.
This may be attributed to the generally slow changes in ocean salinity.
Additionally, we observe that in the salinity forecast, there are instances where a high RMSE is accompanied by a high ACC, as well as cases where a low RMSE is associated with a low ACC.
This is because in the case of an overall high or low salinity field, the spatial distribution of the forecast matches well with the observation, and the model spatial structure anomalies are highly correlated with the observation, but there is an overall systematic bias in the absolute values, resulting in a high RMSE while the ACC is still very high. The overall deviation of the forecast field is very small (mean value, amplitude, etc. are consistent with observations, and the RMSE is small), but the details of the spatial variations (anomaly distributions) are inconsistent with observations, such as the location of the high- and low-value regions is reversed or the phase is reversed, resulting in a small RMSE but also a small ACC. 

\begin{figure*}[!t]
\centering
\includegraphics[width=1\textwidth]{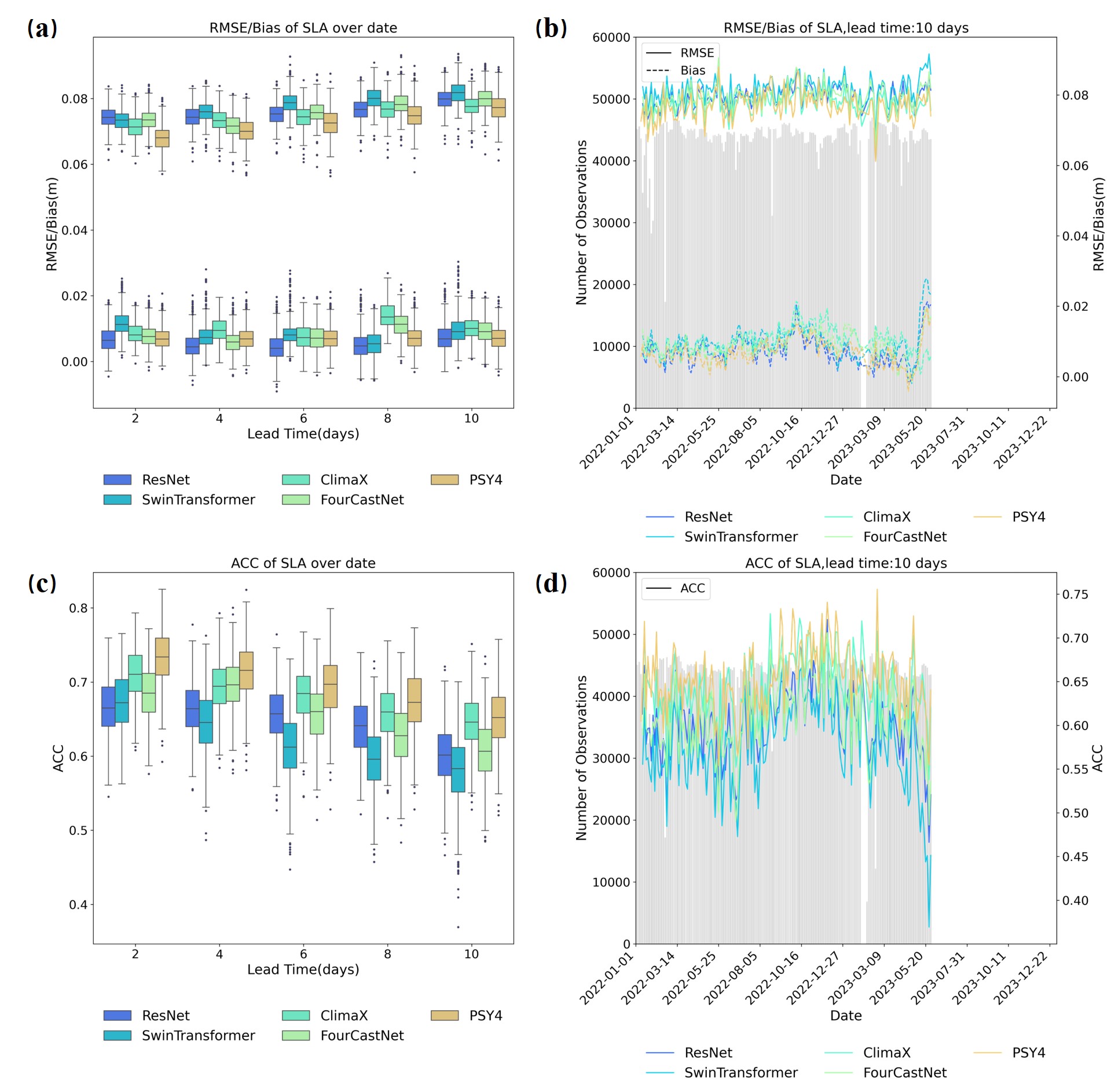} 
\caption{Forecast accuracy against SLA observations taken by CMEMS. (a) RMSE/Bias as a function of forecast lead time. The boxes are the interquartile range and the 75th percentile. (b) RMSE/Bias of 10-day forecasts as a function of date. (c) Similar to (a) with ACC as the metric. (d) Similar to (b) with ACC as the metric.}
\label{sla_date} 
\end{figure*}
\begin{figure*}[!h]
\centering
\includegraphics[width=1\textwidth]{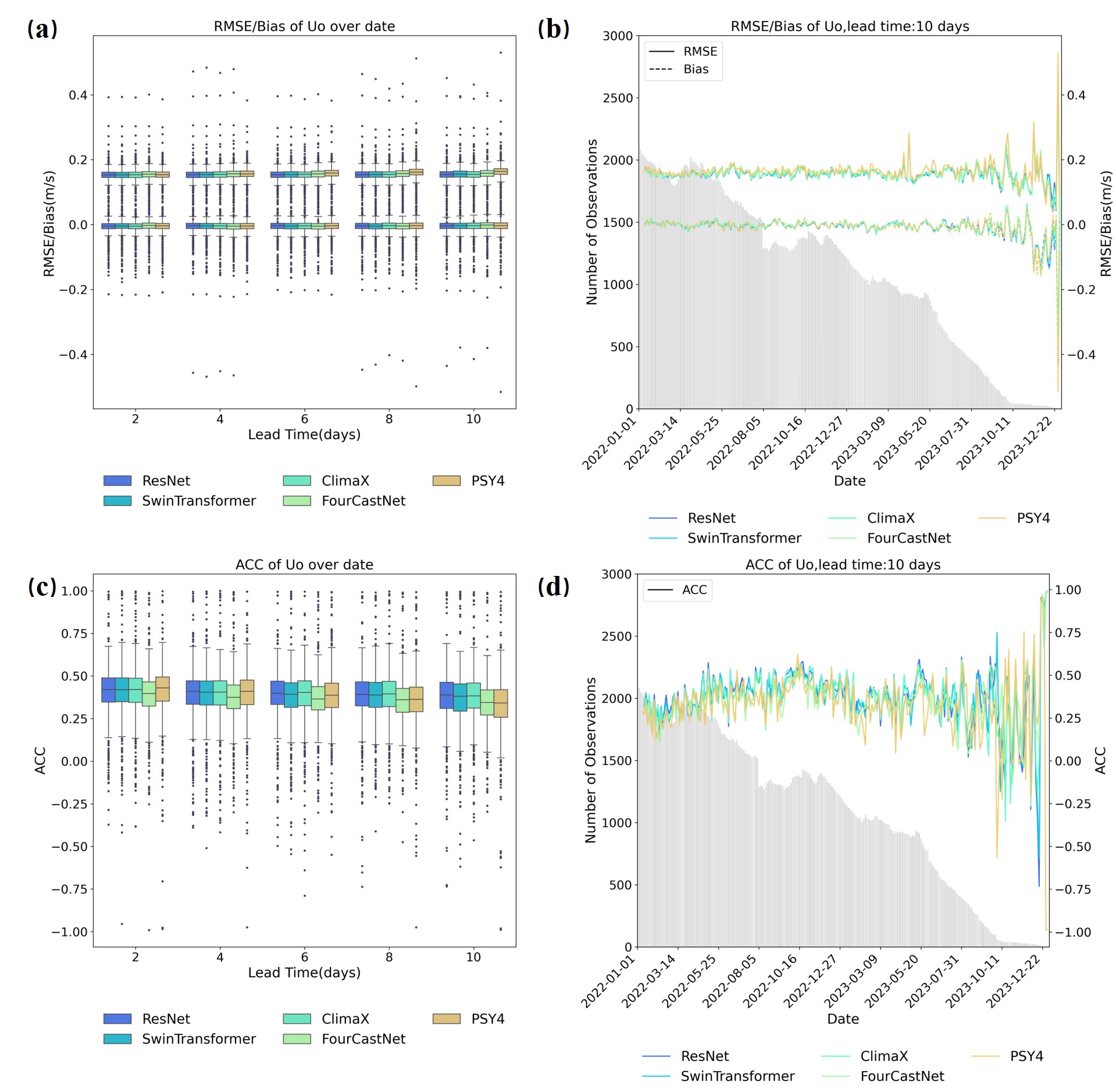} 
\caption{Forecast accuracy against Uo (eastward currents velocity) observations taken by GDP. (a) RMSE/Bias as a function of forecast lead time. The boxes are the interquartile range and the 75th percentile. (b) RMSE/Bias of 10-day forecasts as a function of date. (c) Similar to (a) with ACC as the metric. (d) Similar to (b) with ACC as the metric.}
\label{Uo_date} 
\end{figure*}
\begin{figure*}[!h]
\centering
\includegraphics[width=1\textwidth]{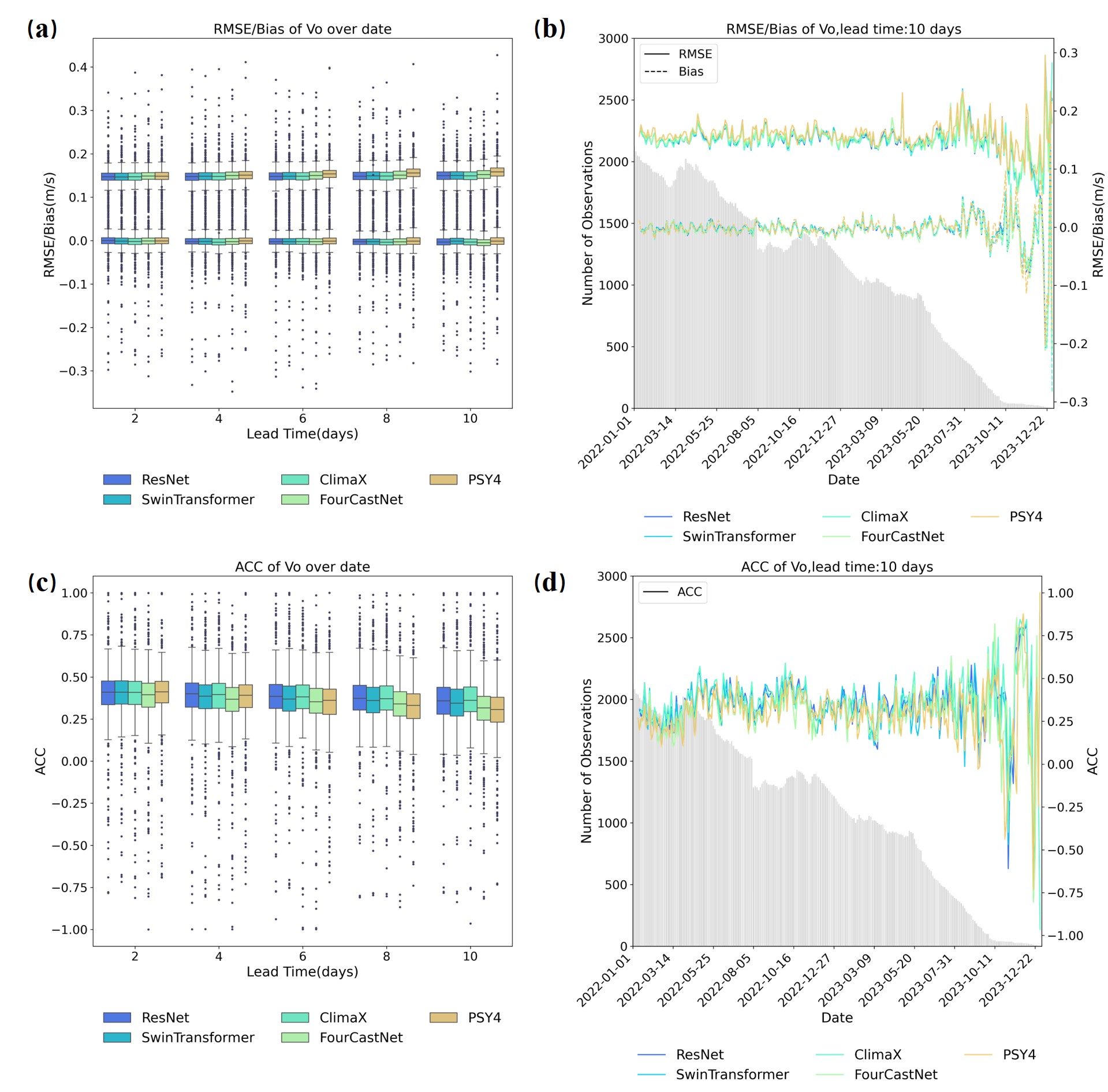} 
\caption{Forecast accuracy against Vo (northward currents velocity) observations taken by GDP. (a) RMSE/Bias as a function of forecast lead time. The boxes are the interquartile range and the 75th percentile. (b) RMSE/Bias of 10-day forecasts as a function of date. (c) Similar to (a) with ACC as the metric. (d) Similar to (b) with ACC as the metric.}
\label{Vo_date} 
\end{figure*}
Figure~\ref{sst_date}a illustrates the RMSE/Bias of SST forecasts as functions of lead time.
Across all lead times, the RMSE values of all models generally increase with longer lead times, indicating the inherent challenge of maintaining forecast accuracy as the lead time extends.
The box plots reveal that model like PSY4 exhibits relatively lower RMSE values in all lead time forecasts, suggesting their superior performance and consistency in SST forecasts.
As shown in Figure~\ref{sst_date}c, the SwinTransformer and FourcastNet models exhibit relatively lower ACC values across the lead time, with their box plots positioned lower and showing larger variations, indicating poorer forecast performance compared to the other models.
From the results of Figure~\ref{sst_date}b and Figure~\ref{sst_date}d, we can observe that after 2023, the RMSE and Bias curves begin to overlap as the number of available GDP buoys decreases.
This overlap signifies significant fluctuations in both metrics, indicating uncertainty in the evaluation results.
As a result, the model's actual forecast performance becomes indistinguishable.
Furthermore, after October 2023, due to a significant reduction in available observational data, the ACC curve exhibited severe fluctuations and showed a negative correlation trend.
Therefore, to ensure that the available observations are sufficient to robustly assess the model performance, the overall performance evaluation of SST was conducted using data from only the year 2022.
This highlights the critical role that the number of in-situ observations plays in effectively assessing the performance of forecasts.

Figure~\ref{sla_date}b presents the RMSE/Bias of 10-day SLA forecasts across different dates.
The RMSE values of all baselines remain relatively stable, but there are noticeable spikes and drops at specific dates, which might be associated with changing oceanic and atmospheric conditions or variations in the quality and quantity of input data.
From the forecast results of SLA in Figure~\ref{sla_date}a, we observed that all models exhibit a positive bias, indicating a consistent tendency to underestimate SLA values. 
As shown in Figure~\ref{sla_date}c, the ACC values of all models generally decrease as the lead time increases, reflecting the diminishing correlation between forecasted and observed SLA over longer prediction periods.

The analysis of Uo and Vo forecast results, as depicted in Figure~\ref{Uo_date} and~\ref{Vo_date}, reveals significant insights into the performance of various models across different lead times.
It can be observed that the RMSE values are distributed within a relatively narrow range, but the ACC values for the velocity forecasts are all below 0.5, with numerous outliers present in all three metrics. This indicates that the performance of all models in currents velocity forecasts is unreliable. A possible reason is the limited and poor-quality velocity data, which increases the difficulty of accurate velocity forecasting.
Other studies indicate that only 3.8\% of the mid-depth ocean (including parts of the equatorial Pacific and the Antarctic Circumpolar Current) can be considered accurately modeled, while significant underestimation of the mean currents velocity is observed in other regions~\cite{su2023widespread}.
The RMSE, Bias, and ACC curves exhibit significant fluctuations after October 2023, and the underlying reasons have been thoroughly analyzed in the SST forecast results.

By evaluating the models' daily forecast performance, the changes in performance over date can be comprehensively displayed. This helps researchers understand the models' generalization ability and robustness.

\begin{figure*}[!h]
\centering
\includegraphics[width=1\textwidth]{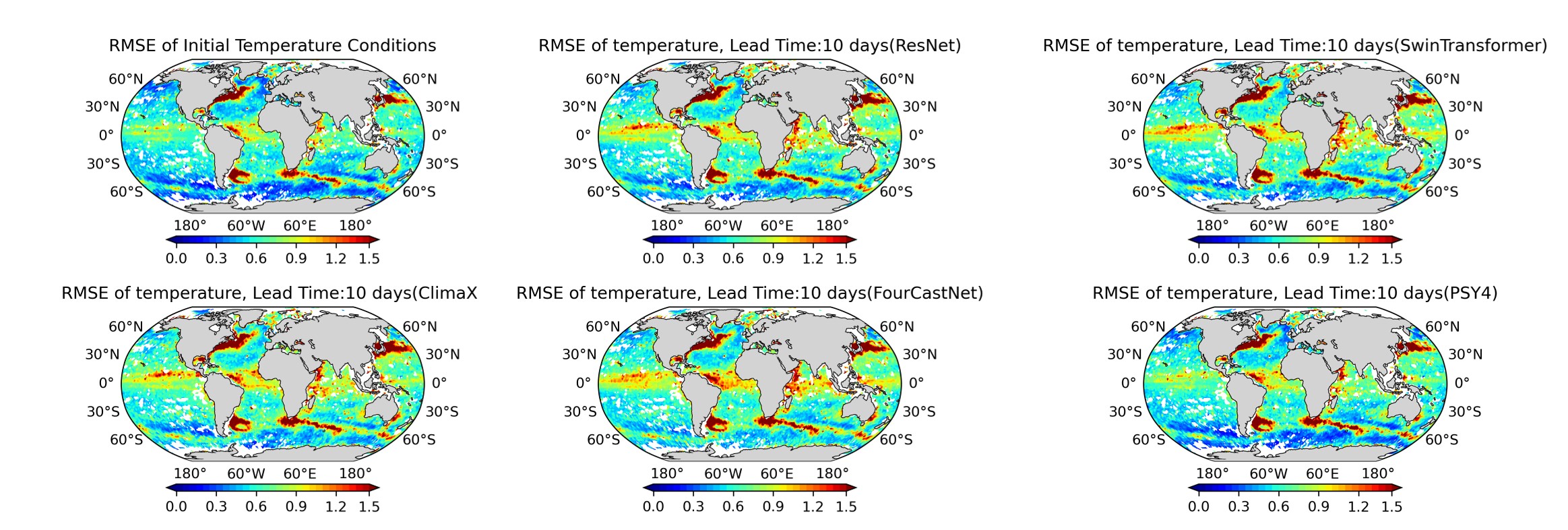}
\caption{Global ocean temperature RMSE (lower is better) distribution map with lead time of 10 days calculated on every 1.40625°×1.40625° area. RMSE is computed against observations taken by EN4.}

\label{temperature_global} 
\end{figure*}

\textbf{Global horizontal forecasting performance.}
The global ocean temperature forecast results, as depicted in Figure~\ref{temperature_global}, reveal the RMSE distribution for various models with a lead time of 10 days.
We observed that all the baseline models exhibit elevated RMSE ($>1.2$) in coastal regions (e.g., western boundary currents and marginal seas) while showing lower errors ($<0.6$) in open tropical oceans ($0^{\circ}-30^{\circ}\text{N/S}$).
The global ocean temperature forecast results demonstrate that the RMSE distributions of all baselines (e.g., ResNet, SwinTransformer, ClimaX, FourCastNet, and PSY4) closely resemble the initial temperature condition errors.
This consistency across models suggests that systematic biases inherent in the initial field data are preserved during the forecasting process, regardless of the algorithmic framework employed.
Coastal regions exhibit significantly higher RMSE values compared to open ocean areas, primarily due to the influence of small- to medium-scale geographical factors. These include nearshore seabed topographical variations, riverine inputs, and localized tidal dynamics, which drive rapid, spatially heterogeneous temperature fluctuations. The RMSE maps reveal concentrated error hotspots along coastlines, reflecting the inability of current models to fully resolve these localized processes.
Although the data-driven approach can provide real-time offshore ocean temperature forecasts, the localized anomalous changes in offshore ocean temperature are significantly influenced by small- and medium-scale geographic factors, and the relative scarcity of observations related to these factors makes it difficult to adequately incorporate the observational information on these influences in the training of the forecast model and in the initial field data. As a result, such data deficiencies usually adversely affect the effectiveness of purely data-driven models in forecasting ocean temperature in offshore regions.

\begin{figure*}[!h]
\centering
\includegraphics[width=1\textwidth]{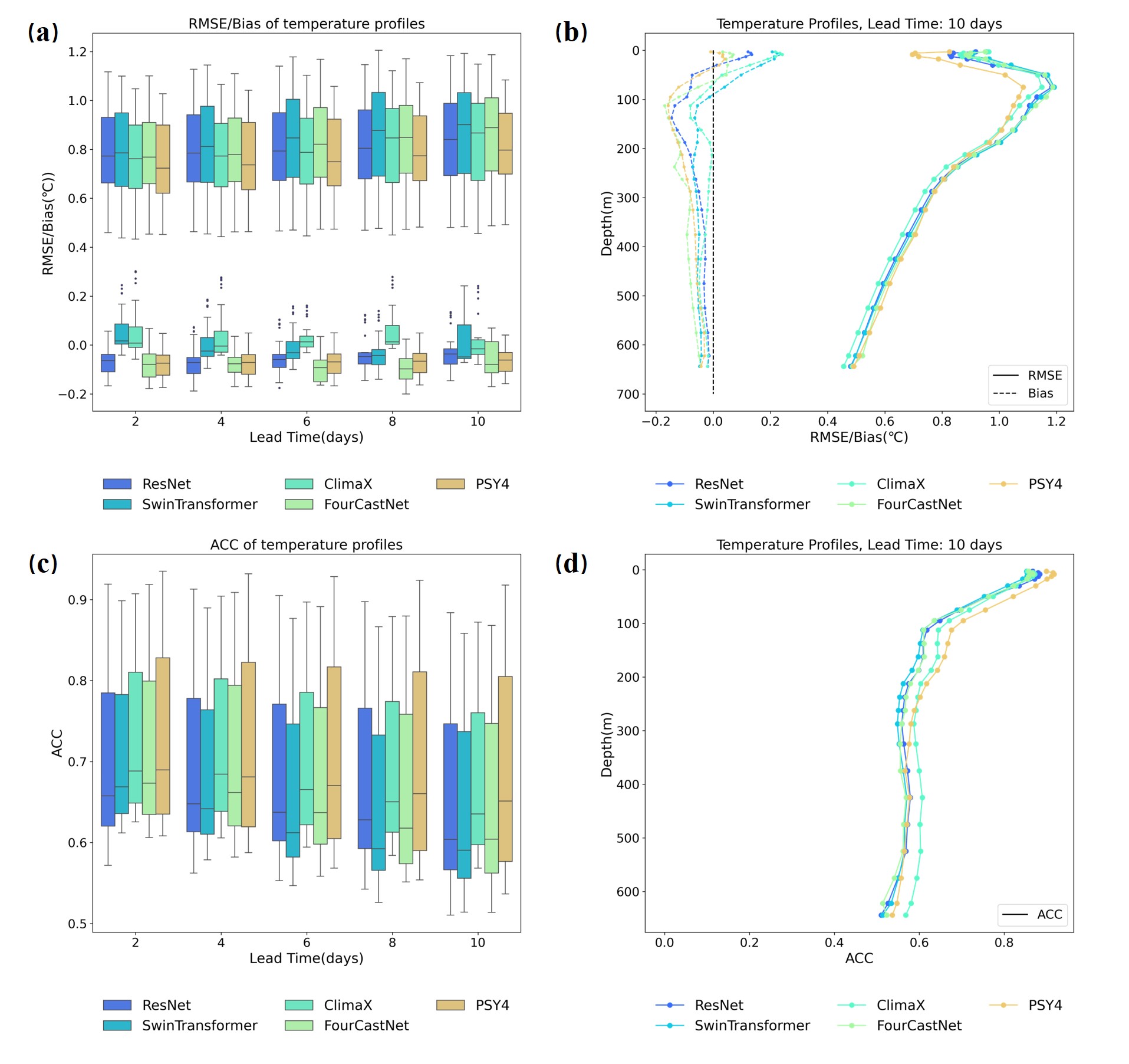}
\caption{Forecast accuracy against temperature profiles taken by EN4. (a) RMSE/Bias as a function of forecast lead time. The boxes are the interquartile range and the 75th percentile. (b) RMSE/Bias of 10-day forecast as a function of depth. (c) Similar to (a) with anomaly correlation as the metric. (d) Similar to (b) but with anomaly correlation as the metric.}
\label{temperature_profiles}
\end{figure*}

\textbf{Vertical forecasting performance.}
The evaluation experiment on the RMSE of temperature over vertical depth was conducted to analyze the accuracy of predicting ocean temperature profiles, which is crucial  for understanding oceanic thermal structures and related processes. 
The analysis of temperature profile forecast results is presented in Figure~\ref{temperature_profiles}, which evaluates the performance of various models against temperature profiles taken by EN4.
As shown in Figure~\ref{temperature_profiles}a, the RMSE/Bias values of all models generally increase as the lead time extends from 1 to 10 days. PSY4 show relatively lower RMSE/Bias values across most lead times, implying better forecast performance.
In contrast, SwinTransformer have relatively higher RMSE/Bias values, indicating larger errors and biases.
Figure~\ref{temperature_profiles}b presents the RMSE and Bias of the 10-day forecast as a function of depth.
As the depth increases from the shallow layers to the thermocline (typically 100–200 meters below the surface), the performance of all models shows a gradual increase in RMSE values, indicating that the baseline models face significant challenges in forecasting the temperature at the depth of the thermocline.
The thermocline is a critical region where temperature changes sharply with depth, and the rapid variation in temperature makes it difficult for models to accurately capture the temperature profile.
The increase in RMSE values further suggests that the baseline models face significant challenges in forecasting the temperature within the thermocline.
Among all models, PSY4 consistently demonstrates the lowest RMSE values across all depth levels, including within the thermocline region. In contrast, ResNet and SwinTransformer exhibit higher RMSE values.
\begin{figure}[!t]
\centering
\includegraphics[width=1\textwidth]{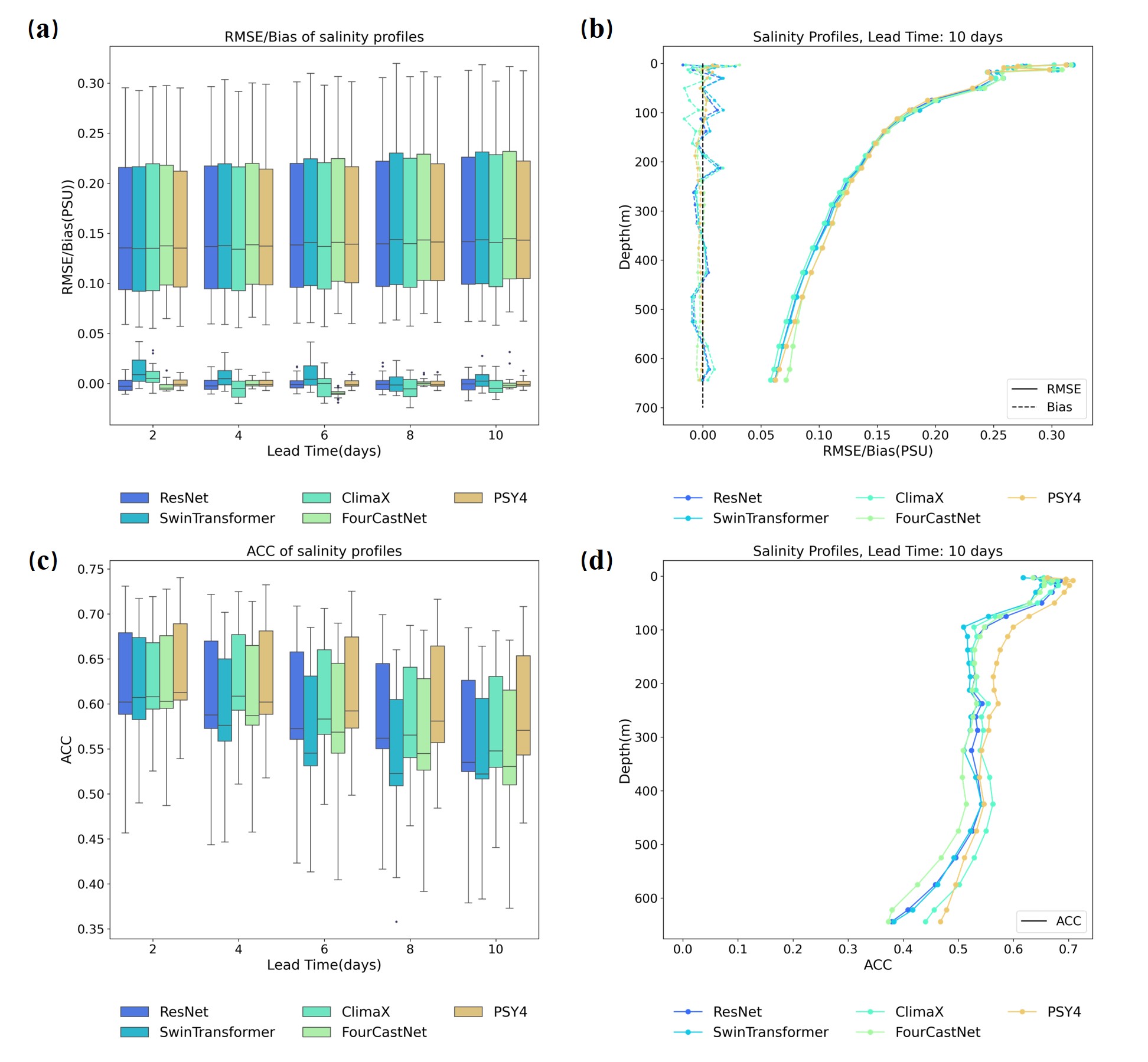}
\caption{Forecast accuracy against salinity profiles taken by EN4. (a) RMSE/Bias as a function of forecast lead time. The boxes are the interquartile range and the 75th percentile. (b) RMSE/Bias of 10-day forecast as a function of depth. (c) Similar to (a) with anomaly correlation as the metric. (d) Similar to (b) but with anomaly correlation as the metric.}
\label{salinity_profiles}
\end{figure}

Figure~\ref{salinity_profiles} presents the forecast accuracy against salinity profiles taken by EN4.
From Figure~\ref{salinity_profiles}a, it can be observed that the RMSE curves of various models show relatively small differences.
However, when it comes to the ACC metric depicted in Figure~\ref{salinity_profiles}c and d, the differences among models become more pronounced.
The ACC measures the correlation between forecasted and observed anomalies, which reflects the model's ability to capture the spatial variations of salinity profiles.
A higher ACC indicates that the model can better forecast the variations in salinity profiles.
The greater differences in ACC curves at different depths once again highlight the varying capabilities of models in representing the vertical structure and variations of salinity profiles.

\textbf{Regional forecasting performance.}
This evaluation experiments were conducted to assess the models' spatiotemporal generalization capabilities across diverse ocean basins, aiming to identify basin-specific performance variations and guide targeted model improvements.
The global ocean forecast results were divided into distinct basins, including the North Atlantic, Tropical Atlantic, South Atlantic, North Pacific, Tropical Pacific, South Pacific, Indian Ocean~\cite{ryan_GODAE_2015}.
The RMSE metrics for six ocean variables were calculated separately for each designated basin.

\begin{figure}
\centering
\includegraphics[width=\textwidth]{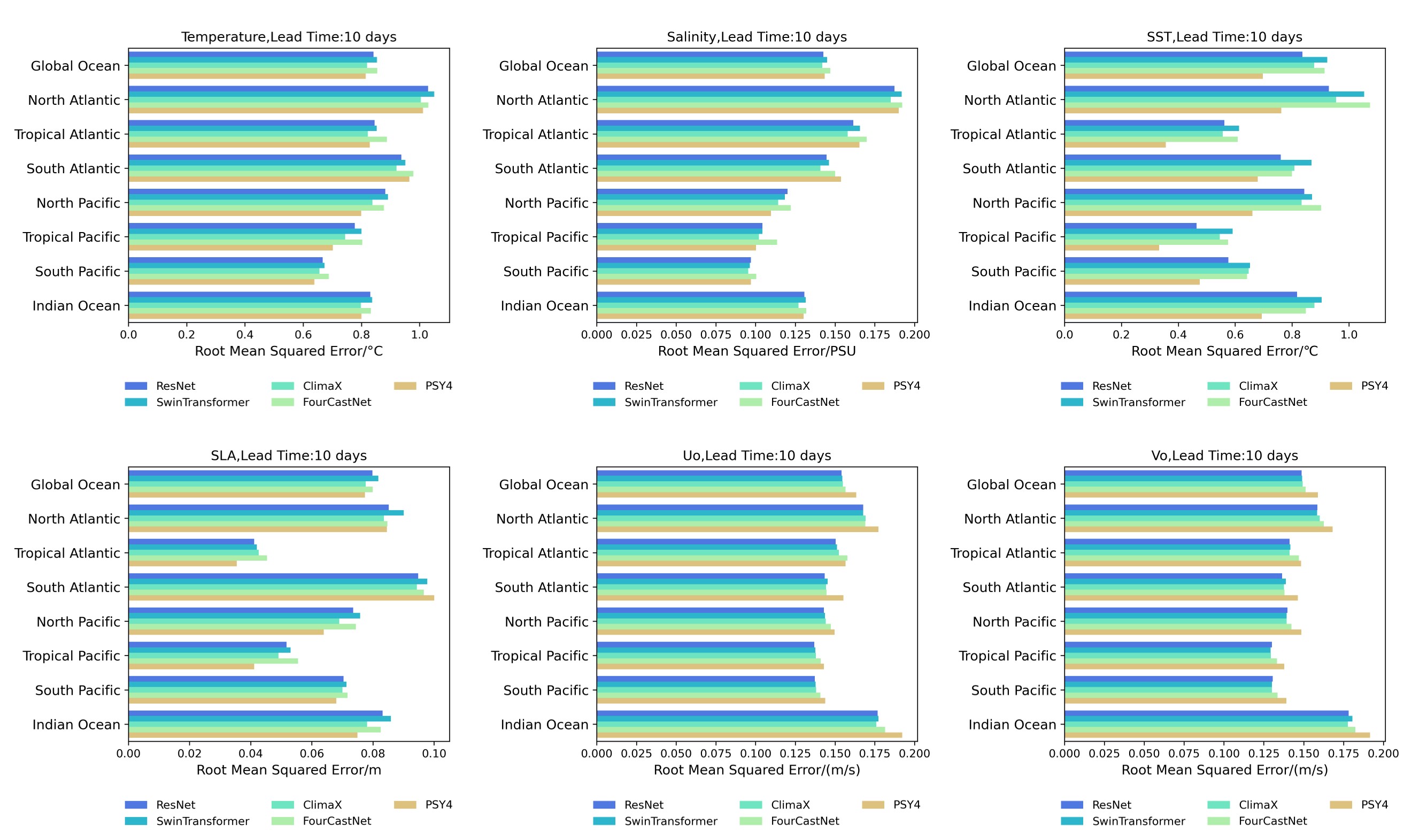}
\caption{The RMSE (lower is better) distribution diagram of different ocean basins with a lead time of 10 days.}
\label{regions}
\end{figure}

The results presented in Figure~\ref{regions} illustrate the RMSE distribution of six ocean variables (temperature, salinity, SST, SLA, Uo and Vo) across different ocean basins with a lead time of 10 days.
For temperature forecasts, the RMSE varies across different basins. The North Atlantic and South Atlantic generally show higher RMSE values compared to other regions, indicating greater challenges in temperature forecasting in these areas. 
In terms of salinity forecasting, the RMSE distribution shows that the North Atlantic and Tropical Atlantic have relatively higher errors. This might be associated with the complex oceanic processes and variability in these regions. 
The SST forecast results show a different trend to temperature, with the PSY4 performing best overall.
The differences in RMSE values across different ocean basins are significant. For example, there is a notable disparity in SST forecast performance between the Tropical Pacific and the North Atlantic.
Regarding SLA, the RMSE tends to be higher in the South Atlantic and North Atlantic.
The performance of different models in SLA forecasting varies, with PSY4 achieving relatively lower RMSE values across several basins.
In the case of Uo and Vo, the RMSE distribution indicates that forecasting ocean currents remains challenging, especially in the Indian Ocean. 
The ResNet shows some superiority in currents velocity forecasting, with lower RMSE values in multiple ocean basins.

\section{Conclusion}

We introduce OceanForecastBench, an open-source benchmark for training and evaluating data-driven models to forecast key ocean variables, including SLA, SST, temperature, salinity, and currents.
For model training, OceanForecastBench provides a standardized training dataset with consistent spatial and temporal resolution, spanning the past 28 years. 
This dataset is derived from over 13 TB of raw reanalysis data, with GLORYS12 serving as the primary source. 
To improve the representation of air-sea interactions,  the training set also incorporates sea surface wind fields from ERA5 and SST data from OSTIA.
For model evaluation, OceanForecastBench provides a robust evaluation dataset containing over 100 million observations collected from satellite remote sensing and in-situ measurements. 
A standardized evaluation pipeline is developed to align forecast outputs with observational data in spatialtemporal dimension and to compute performance metrics in a consistent manner.
This benchmark evaluate 5 baseline models, encompassing both operational numerical forecasting systems and advanced deep learning models, offering an in-depth comparison of their capabilities. 
The results highlight the respective strengths and limitations of each model, offering guidance for future methodological improvements.
OceanForecastBench standardizes both the training and evaluation processes, simplifies data processing workflows, and lowers the barrier for interdisciplinary engagement in data-driven ocean forecasting research.
This benchmark provides a foundation for future developments in ocean modeling and forecasting, offering a valuable resource for advancing the understanding and forecasting of oceanic phenomena.

\section*{Data Availability Statement}
The GLORYS12 reanalysis can be found at \url{https://data.marine.copernicus.eu/product/GLOBAL_MULTIYEAR_PHY_001_030/services}~\cite{GLORYS12}, the sea surface wind data from the Fifth Generation Global Atmospheric Reanalysis (ERA5)19 at \url{https://cds.climate.copernicus.eu/cdsapp#!/dataset/reanalysis-era5-single-levels?tab=form}~\cite{ERA5DATA} and SST data from the Operational Sea Surface Temperature and Ice Analysis (OSTIA) at \url{https://data.marine.copernicus.eu/product/SST_GLO_SST_L4_REP_OBSERVATIONS_010_011/services}~\cite{good2020current}. We obtained the PSY4 global physics analysis and forecast products from January 1, 2022, to December 31, 2023 from the link \url{https://data.marine.copernicus.eu/product/GLOBAL_ANALYSISFORECAST_PHY_001_024/services}~\cite{analysis}. Furthermore, the EN4 data can be downloaded at \url{https://hadleyserver.metoffice.gov.uk/en4/download-en4-2-2.html}~\cite{good2013en4}, GDP data at \url{https://www.aoml.noaa.gov/phod/gdp/}~\cite{lumpkin2019global}
, and CMEMS L3 data at \url{https://data.marine.copernicus.eu/product/GLOBAL_ANALYSISFORECAST_PHY_001_024/services}~\cite{cmemsl3}

\acknowledgments
This research is partially supported by National Key R\&D Program of China (2024YFC3109200), Hunan Provincial Natural Science Foundation of China (2024JJ4042), the science and technology innovation Program of Hunan Province (2024RC3134), and Youth Independent Innovation Science Fund of the National University of Defense Technology (ZK24-53).

The EN.4.2.2 data used by this work to construct evaluation dataset were obtained from \url{https://www.metoffice.gov.uk/hadobs/en4/} and are © British Crown Copyright, Met Office, [2025], provided under a Non-Commercial Government Licence \url{http://www.nationalarchives.gov.uk/doc/non-commercial-government-licence/version/2/}.

This study has been conducted using E.U.Copernicus Marine Service Information, \url{https://doi.org/10.48670/moi-00021}, \url{https://doi.org/10.48670/moi-00168}, \url{https://doi.org/10.48670/moi-00146}.

\bibliography{reference_full}

\end{document}